\documentclass[conference]{IEEEtran}
\IEEEoverridecommandlockouts

\usepackage{cite}
\usepackage{amsmath,amssymb,amsfonts}
\usepackage{graphicx}
\usepackage{textcomp}
\usepackage{xspace}
\usepackage{xcolor}
\usepackage{booktabs} 
\usepackage{amsmath}
\usepackage{amsfonts}
\usepackage{amssymb}
\usepackage{comment}
\usepackage{footmisc}
\usepackage{mathrsfs}
\usepackage{graphicx}
\usepackage{subcaption}
\usepackage{mwe}
\usepackage{multirow}
\usepackage{algorithm}
\usepackage{algorithmicx}
\usepackage{algpseudocode}
\usepackage{multicol}
\usepackage{enumitem}
\usepackage{array}
\usepackage{float}
\usepackage{hyperref}
\usepackage{cleveref}
\usepackage{bbold}

\def\BibTeX{{\rm B\kern-.05em{\sc i\kern-.025em b}\kern-.08em
    T\kern-.1667em\lower.7ex\hbox{E}\kern-.125emX}}

\crefname{equation}{}{}
\Crefname{equation}{Equ}{Equations}

\newcommand{\best}[1]{$\bf{#1}^{\ast}$}

\newcommand{\eg}{\emph{e.g.,}\xspace}
\newcommand{\ie}{\emph{i.e.,}\xspace}
\newcommand{\etc}{\emph{etc.}\xspace}

\begin{document}

\title{Adversarial Batch Inverse Reinforcement Learning: Learn to Reward from Imperfect Demonstration for Interactive Recommendation}

\author{
\IEEEauthorblockN{Jialin Liu\textsuperscript{1,2}, Xinyan Su\textsuperscript{1,2}, Zeyu He\textsuperscript{3}, Xiangyu Zhao\textsuperscript{4}, Jun Li\textsuperscript{1}}
\IEEEauthorblockA{
{\textsuperscript{1} Computer Network Information Center, Chinese Academy of Sciences, Beijing, China}\\
{\textsuperscript{2} University of Chinese Academy of Sciences, Beijing, China}\\
{\textsuperscript{3} Beijing Information Science and Technology University, Beijing, China}\\
{\textsuperscript{4} City University of Hong Kong, Hong Kong, China}\\
{jlliu@cnic.cn, suxinyan@cnic.cn, hezeyu222638@bistu.edu.cn, xy.zhao@cityu.edu.hk, lijun@cnic.cn}}
}

\maketitle

\begin{abstract}
Rewards serve as a measure of user satisfaction and act as a limiting factor in interactive recommender systems. In this research, we focus on the problem of learning to reward (LTR), which is fundamental to reinforcement learning. Previous approaches either introduce additional procedures for learning to reward, thereby increasing the complexity of optimization, or assume that user-agent interactions provide perfect demonstrations, which is not feasible in practice. Ideally, we aim to employ a unified approach that optimizes both the reward and policy using compositional demonstrations. However, this requirement presents a challenge since rewards inherently quantify user feedback on-policy, while recommender agents approximate off-policy future cumulative valuation. To tackle this challenge, we propose a novel batch inverse reinforcement learning paradigm that achieves the desired properties. Our method utilizes discounted stationary distribution correction to combine LTR and recommender agent evaluation. To fulfill the compositional requirement, we incorporate the concept of pessimism through conservation. Specifically, we modify the vanilla correction using Bellman transformation and enforce KL regularization to constrain consecutive policy updates. We use two real-world datasets which represent two compositional coverage to conduct empirical studies, the results also show that the proposed method relatively improves both effectiveness (2.3\%) and efficiency (11.53\%).
\end{abstract}


\section{introduction}

Modern recommendation technology changes human-machine collaboration from machine-centric searching to human-oriented mining \cite{gao2023cirs},  thus widely accelerating applications like e-commerce \cite{10.1145/3219819.3219886}, \etc From system perspective, a recommender agent mines personalization from user-agent interactions. As these interactions accumulate chronically, the agent gradually learns to imitate user preference and narrows recommendation down to relevant choices that maximize user satisfaction \cite{10.1145/3397271.3401174}. Recently, advancement of reinforcement learning (RL) offers new kits to model this maximization procedure as an interactive decision making process, known as interactive recommender system (IRS), as both on-the-spot rewards from previous behavior demonstrations and off-the-spot rewards from future long-term accumulation are valuable \cite{xiao2021general}.


Reward function quantifies user satisfaction in RL, thus learning to reward (LTR) is fundamental \cite{fu2018learning}. Philosophically, LTR reflects the ability of introspection, a human-level intelligence researchers have pursued. Computationally, rewards transform user satisfaction maximization into discounted future reward cumulation, making it a bottleneck for IRS. However, LTR is challenging: (i). \textbf{Reward equivalence}: multiple reward settings can interpret the optimal recommending policy from the demonstration dataset, making LTR underdetermined\cite{ng1999policy}; (ii). \textbf{Exploration-efficiency} \cite{jing2020reinforcement}: common RL approaches acquire online interaction for policy evaluation, such on-policy planning is constrained in recommendation since under-optimized agents may hurt user satisfaction \cite{schnabel2018short}, thus more sample-efficient approach is acquired. 

Previous works primarily address reward equivalence by employing a separate procedure to approximate rewards based on multiple feedback signals(\eg click, purchase, \etc): Non-adversarial approximations \cite{gong2019exact,xian2019reinforcement} learn a heuristic reward with neural architectures representing inductive bias. Comparatively, adversarial approximation methods learn a discriminating score between demonstration data and recommender agents, when proceeding to the RL planning, this discriminating reward encourages actions similar to behavior patterns. Recently, observing that high-quality demonstrations collected via unknown-yet-unrandom behavior agents are available, several batch RL methods leverage off-policy correction for exploration-efficiency \cite{chen2019top,chen2019generative}. However, existing RL methods for reward-equivalence and exploration-efficiency do not mutually benefit from each other, recent works \cite{chen2019generative,kostrikov2019imitation} aim to bridge the gap, while either still requiring an individual procedure to optimize or assuming data coverage. 
Learn to reward is challenging. First, joint optimization is desired yet contradictory in hitherto methods, either adversarial or supervised learnt reward in its nature is an immediate credit quantification of user feedback, and requires on-policy update. Second, it is relatively straightforward to define the cumulative rewards episodically \cite{bai2019model} rather than immediately  \cite{chen2019generative}, immediate credit assignment is more burdensome \cite{xin2020self}. Additionally, there are two commonly adopted disciplines for offline environments: imitation learning \cite{kostrikov2019imitation} which converges to an implicit reward with expert demonstrations (perfect coverage), and vanilla batch RL \cite{prudencio2023survey} which generally approximates an explicit reward from more random demonstrations (uniform coverage) \cite{rashidinejad2021bridging}. However, compositional demonstration sets (imperfect coverage) between these two extremes are more practical, whose quality is guaranteed by unknown prior behavior agents and thus is uncheckable.

To address the aforementioned challenges, we propose a novel adversarial batch reinforcement learning method for IRS. We utilize discounted stationary distribution correction to combine LTR and policy REINFORCE without requiring additional pipelines.The Bellman transformation on immediate rewards turns on-policy objective into off-policy procedures. For imperfection demonstration, we leverage KL conservation as a form of pessimism \cite{rashidinejad2021bridging} to balance exploitation and exploration. Our main contributions are as follows:
\begin{itemize}
    \item For the first time, we propose a batch inverse RL method for the interactive recommender system with imperfection concerned. It reduces additional learning pipeline for LTR and adapts to different compound demonstrations.
    \item Our conservative learning objective relatively improves 2.3\% over the second best comparison with an 11.53\% reduction on demonstration consumption.
    \item Empirical studies on two real-world recommendation datasets that represent two compositional coverage also demonstrate the effectiveness of our method.
\end{itemize}
\section{problem statement}
Interactions between recommender agent and users can be modeled as Markov Decision Process $(\mathcal{S}, \mathcal{A}, \mathcal{P}, \mathcal{R}, \gamma)$:
\begin{itemize}
    \item \textbf{State space} $\mathcal{S} \in \mathbb{R}^{d_s}$: State $\mathbf{s}$ represents browsing history so far, with each item in the browsing window sorted chronologically to learn state representation.
    \item \textbf{Action space} $\mathcal{A} \in \mathbb{R}^{N}$: The action at time $t$ represents an item back to a user. Without loss of generalization, we assume that the agent will return one item at each time, and list-wise extension is straightforward.
    \item \textbf{Reward} $\mathcal{R}: \mathcal{S} \times \mathcal{A} \rightarrow \mathbb{R}$: The user $\mathbf{s}_t$ browsers received recommendation $\mathbf{a}_t$ at this time he can skip, click or purchase the recommendation. Then the agent receives an immediate reward $r(\mathbf{s}_t, \mathbf{a}_t)$ quantifying the user feedback.
    \item \textbf{Transition probability} $\mathcal{P}: \mathcal{S} \times \mathcal{A} \times \mathcal{S} \rightarrow \mathbb{R}$: probability $p(\mathbf{s}_{t+1} | \mathbf{s}_t, \mathbf{a}_t)$ defines the user state transmission at state $\mathbf{s}_t$ after receive recommendation $\mathbf{a}_t$. We assume this transition satisfies the first order Markov property $p\left(\mathbf{s}_{t+1} \mid \mathbf{s}_t, \mathbf{a}_t, \ldots, \mathbf{s}_1, \mathbf{a}_1\right)=p\left(\mathbf{s}_{t+1} \mid \mathbf{s}_t, \mathbf{a}_t\right)$.
    \item \textbf{Discount factor} $\gamma \in [0, 1]$: $\gamma$ characterizes the importance of different timestamps. Specifically, $\gamma=0$ only values immediate feedback, and $\gamma=1$ will equally contribute all future reward in interactions.
\end{itemize}
For hitherto interaction IDs, the recommender agent $\pi_\theta(\mathbf{a} \mid \mathbf{s})$ uses constructed demonstrations $\mathcal{D} = \{(\mathbf{s}_t, \mathbf{a}_t, \mathbf{s}_{t+1})\}_{k=1}^N$ to maximize following user satisfaction:
\begin{equation} \label{eq:rl}
    \max _{\pi_\theta} \mathbb{E}_{\tau \sim \pi_\theta}\left[\sum_{t=0}^{|\tau|} \gamma^t r\left(\mathbf{s}_t, \mathbf{a}_t\right)\right],
\end{equation}
where $\tau=\left(\mathbf{s}_0, \mathbf{a}_0, \dots, \mathbf{s}_{|\tau|-1}, \mathbf{a}_{|\tau|-1}\right)$ represents an episodic interaction between the agent and users. 
\section{methodology}
Current methods either require divided optimization or assume coverage of demonstration sets, both of which are impractical. In this section, we introduce a novel inverse reinforcement learning paradigm based on discounted stationary distribution correction. We implement Bellman transformation on stationary-action valuation so that the vanilla learning objective is off-policy. And we utilize KL conservation as a pessimism to handle compositional coverage. Finally, we introduce an extensible neural architecture for optimization.

\subsection{Adversarial Inverse Reinforcement Learning} 
 When learning from demonstrations $\mathcal{D}$, reward $r(\mathbf{s}_t, \mathbf{a}_t)$ guides the recommender agent $\pi_\theta(\mathbf{a}\mid\mathbf{s})$ towards unknown behavior agents which collects $\mathcal{D}$:
\begin{equation} \label{eq:dsd}
    \begin{aligned}
    r\left(\mathbf{s}_t, \mathbf{a}_t\right) & =\log \frac{d^{\mathcal{D}}\left(\mathbf{s}_t, \mathbf{a}_t\right)}{d^{\pi_\theta}\left(\mathbf{s}_t, \mathbf{a}_t\right)},
    \end{aligned}
\end{equation}
where $d^{\pi_\theta}(\mathbf{s}_t, \mathbf{a}_t) \varpropto (1-\gamma)\sum_{t=0}\gamma^t exp\left( h_{\theta}(\mathbf{s}_t, \mathbf{a}_t)\right)$ is parameterized discounted stationary distribution \cite{kostrikov2019imitation}, and it induces discounted factor $\gamma$ to tackle distribution shift \cite{ross2010efficient}. Vanilla RL objective \cref{eq:rl} then transforms into:
\begin{equation}\label{eq:ddm}
\max _{\pi_\theta}(1-\gamma) \cdot \mathbb{E}_{\left(\mathbf{s}_t, \mathbf{a}_t) \sim \mathcal{D}\right.}\left[\sum_{t=0}^{\infty} \gamma^t \log \frac{d^{\mathcal{D}}\left(\mathbf{s}_t, \mathbf{a}_t\right)}{d^{\pi_\theta}\left(\mathbf{s}_t, \mathbf{a}_t\right)}\right],
\end{equation}
which can be further expanded as \cite{kostrikov2019imitation}:
\begin{equation}
    \label{eq:dice_part1}
    \max_{\pi_\theta} \log \mathbb{E}_{(\mathbf{s}, \mathbf{a}) \sim d^{\mathcal{D}}}\left[e^{r_\phi(\mathbf{s}, \mathbf{a})}\right]-(1-\gamma)\mathbb{E}_{(\mathbf{s}, \mathbf{a}) \sim d^{\pi_\theta}}\left[r_\phi(\mathbf{s}, \mathbf{a})\right].
\end{equation}
Although the quality of $\mathcal{D}$ is unknown in prior, reward offers valuation information that the agent $\pi_\theta$ can later use to reformulate new transitions which has not been yet observed in $\mathcal{D}$. To learn $r_\phi(\mathbf{s}, \mathbf{a})$, we imitate the behavior cumulative valuation in a min-max game which converges to \cref{eq:dsd}:
\begin{equation}
    \label{eq:dice_part2}
    \begin{aligned}
    \max_{\pi_\theta} \min_{r_\phi} & \log \mathbb{E}_{(\mathbf{s}, \mathbf{a}) \sim d^{\mathcal{D}}}\left[e^{r_\phi(\mathbf{s}, \mathbf{a})}\right] \\
    - & (1-\gamma)\mathbb{E}_{(\mathbf{s}, \mathbf{a}) \sim d^{\pi_\theta}}\left[r_\phi(\mathbf{s}, \mathbf{a})\right],
    \end{aligned}
\end{equation}
For the minimization part, parameterized reward $r_\phi(s,a)$ tries to imitate valuation pattern in demonstration $\mathcal{D}$; For the maximization part, the agent $\pi_\theta$ generally reformulates new sub-pattern from existing interaction episodes.

\subsection{Bellman Transformation for Efficiency} 
The on-policy evaluation part from $d^{\pi_\theta}$ in \cref{eq:dice_part2} leads to low efficiency. We utilize Bellman operator as,
\begin{equation}
    \label{eq:bellman}
    \mathcal{B}^\pi v(\mathbf{s}, \mathbf{a})=\gamma \mathbb{E}_{\mathbf{s}' \sim p(\cdot \mid \mathbf{s}, \mathbf{a}), \mathbf{a}' \sim \pi\left(\cdot \mid \mathbf{s}'\right)}\left[v\left(\mathbf{s}', \mathbf{a}'\right)\right],
\end{equation}
where state $\mathbf{s}'$ and action $\mathbf{a}'$ is the next timestamp of state $\mathbf{s}$, here we temporarily drop time subscripts to represent a general triple $(\mathbf{s}, \mathbf{a}, \mathbf{s}')$. Reward \cref{eq:dice_part2} is then computed as a temporal difference between consecutive tuples:
\begin{equation}
    \label{eq:transform}
    r_\phi(\mathbf{s}, \mathbf{a}) = v_\phi(\mathbf{s}, \mathbf{a}) - \mathcal{B}^{\pi}v_\phi(\mathbf{s}, \mathbf{a}),
\end{equation}
where $v_\phi(\mathbf{s}, \mathbf{a})$ is a cumulative state-action valuation approximation (parameterized by $\phi$). Combined with Bellman transformation \cref{eq:transform}, the on-policy part \cref{eq:dice_part2} reduces into a linear form  which leads to an off-policy version:
\begin{equation}
    \label{eq:dice}
    \begin{aligned}
    \max_{\pi_\theta} \min_{v_\phi} & \log \mathbb{E}_{(\mathbf{s}, \mathbf{a}, \mathbf{s}') \sim d^{\mathcal{D}}}\left[e^{r_\phi(\mathbf{s}, \mathbf{a})}\right] \\
    & - (1-\gamma)\mathbb{E}_{(\mathbf{s}_0, \mathbf{a}_0) \sim d^{\mathcal{D}}}\left[r_\phi(\mathbf{s}_0, \mathbf{a}_0)\right].
    \end{aligned}
\end{equation}
This objective exhibits two characteristics which are absent in previous work: Firstly, it does not acquire another separated training pipeline for LTR, thereby avoiding additional complexity; Secondly, it does not require on-policy interactions from users, and thus improves the efficiency.

\subsection{KL Conservation for Effectiveness} 
One problem of the vanilla objective \cref{eq:dice} is that it purely relies on a demonstration set. In practice, the quality of $\mathcal{D}$ can be compositional between perfect coverage and uniform coverage, which goes beyond the original data assumptions. Furthermore, the demonstration sets may lack diversity. Inspired by the concept of pessimism as an inductive bias in risky complex environments \cite{rashidinejad2021bridging}, we restrain consecutive updates  within a divergence measure via:
\begin{equation}
    \label{eq:cdice}
    \begin{aligned}
        \max_{\pi_\theta} \min_{v_\phi} & \log \mathbb{E}_{(\mathbf{s}, \mathbf{a}, \mathbf{s}') \sim d^{\mathcal{D}}}\left[e^{r_\phi(\mathbf{s}, \mathbf{a})}\right] \\
        & - (1-\gamma)\mathbb{E}_{(\mathbf{s}_0, \mathbf{a}_0) \sim d^{\mathcal{D}}}\left[r_\phi(\mathbf{s}_0, \mathbf{a}_0)\right] \\
        & - \mathbb{KL}\left[\pi_{\theta}(\cdot | \mathbf{s}) || \pi_{\theta'}(\cdot | \mathbf{s})\right].
    \end{aligned}
\end{equation}
The KL penalty can be approximated by the Fisher information matrix $G(\cdot;\theta)$ \cite{kakade2001natural} with the second-order Taylor expansion. Thus we achieve the overall adversarial batch inverse reinforcement learning objective. From optimization perspective, vanilla objective \cref{eq:dice} unifies LTR in the minimization and policy REINFORCE \cite{chen2019top} in the maximization, KL regularization removes uncertainty and takes conservative gradient steps. This is different from mixture regularization which still acquires online interaction for diversity \cite{kostrikov2019imitation}.

\subsection{Neural Implementation}
In order to estimate the conservative objective \cref{eq:cdice}, we adopt an extensible Actor-Critic architecture, which consists of two components: (i). the \textbf{critic} $v_\phi(\mathbf{s}, \mathbf{a})$ that valuates the reward implicitly with off-policy demonstrations. (ii). the \textbf{actor} $\pi_\theta(\mathbf{a} \mid \mathbf{s})$ that generates recommendation based on its policy. Both components share the same state encoding backbone, which forms a simplified mixture of experts.

\noindent \textbf{Encoder} Given a demonstration $\{i_0, i_1, \dots, i_{t-1}\}$, the encoder first projects recorded item $i_{t-1}$ into an embedding vector $\mathbf{e}_{t-1} \in \mathbb{R}^{d_e}$. We then use autoregressive neural networks to model transition probability $p(\mathbf{s}_t | \mathbf{s}_{t-1}, \mathbf{a}_{t-1})$, and the state $\mathbf{s}_t$ can be formalized as follows:
\begin{equation}
    \label{eq:encoder}
    \mathbf{s}_t = h_{\theta_e}(\mathbf{s}_{t-1}, \mathbf{e}_{t-1}),
\end{equation}
where $\theta_e$ denotes learnable parameters, and the autoregressive model $h_{\theta_e}(\cdot, \cdot)$ can be recurrent neural network, \ie GRU \cite{hidasi2018recurrent} or feedforward neural network, \ie CNN \cite{tang2018personalized}. For both architectures to capture temporal dynamics of transitions, we use a $w$-length window and concatenate the recent interactions $[i_{t-w+2}, i_{t-w+3}, \dots, i_{t-1}]$ sampled from $\mathcal{D}$, with truncating (if $t > w$) and padding at the rightmost (if $t < w$).

\noindent \textbf{Actor} Based on current state $\mathbf{s}_t$, the actor agent generates a list of candidates from the entire item space $|\mathcal{A}|$ as its next action $\mathbf{a}_t$. A straightforward representation of item $i$ to be involved under current user state $\mathbf{s}_t$ is thus:
\begin{equation}
    \label{eq:vanilla-policy}
    \pi\left(i \in \mathbf{a}_t \mid \mathbf{s}_t\right)=\frac{\exp \left(\mathbf{W}_i \mathbf{s}_t+\mathbf{b}_i\right)}{\sum_{j=1}^{|\mathcal{A}|} \exp \left(\mathbf{W}_j \mathbf{s}_t+\mathbf{b}_j\right)},
\end{equation}
where $\mathbf{W}_i$ is the $i$-th row of parametric matrix $\mathbf{W}_{(a)} \in \mathbb{R}^{|\mathcal{A}| \times d_s}$, $\mathbf{b}_{(a)} \in \mathbb{R}^{d_s}$ is the corresponding bias. Due to the large action space in recommendation ($|\mathcal{A}| \gg 1$), vanilla policy \cref{eq:vanilla-policy} is expensive to enumerate. We utilize the Gumbel-Softmax trick to provide a differentiable approximation:
\begin{equation}
    \label{eq:gm-policy}
    \pi\left(i \in \mathbf{a}_t \mid \mathbf{s}_t\right) = \frac{\exp \left(\left(\log \left(f_{\theta_a}\left(\mathbf{s}_t\right)\left[\mathbf{e}_i\right]\right)+g_i\right) / \gamma_g\right)}{\sum_{j=1}^{|\mathcal{A}|} \exp \left(\left(\log \left(f_{\theta_a}\left(\mathbf{s}_t\right)\left[\mathbf{e}_j\right]\right)+g_j\right) / \gamma_g\right)},
\end{equation}
where $\{g_j\}_{j=1}^{|\mathcal{A}|}$ is i.i.d. samples from Gumbel distribution, $\gamma_g$ is the scalar temperature, $f_{\theta_a}$ is a multi-layer perceptron which maps user current state into action preferences. \Cref{eq:gm-policy} replace the argmax with discrete softmax, such replacement can avoid distribution mismatch \cite{xiao2021general}.

\noindent \textbf{Critic} To implicitly learn reward from the compositional demonstration, we take concatenation of current state $\mathbf{s}_t$ and potential action (item) $\mathbf{e}_t^{\mathbf{(a)}}$ as the input of the critic $v_\phi(\mathbf{s}, \mathbf{a})$, which measures discounted cumulative rewards as:

\begin{equation}
    \label{eq:critic}
    v_\phi(\mathbf{s}, \mathbf{a}) = \mathbf{w}_{(c)}^T\sigma\left(\mathbf{W}_{(c)}\left[(\mathbf{s}_t)^T, (\mathbf{e}_{t}^{\mathbf{(a)}})^T\right]^T + \mathbf{b}_{(c)}\right),
\end{equation}

where $\mathbf{W}_{(c)}\in \mathbb{R}^{l \times (d_s + d_e)}$ is the weight matrix,  $\mathbf{b}_{(c)} \in \mathbb{R}^{l}$ denotes the bias term, and $\mathbf{w}_{(c)} \in \mathbb{R}^{l}$ is the regression parameters. $\sigma(\cdot)$ is the nonlinear activation such as ReLU. For anotating simplicity, we resemble $\phi = \{\mathbf{w}_{(c)}, \mathbf{W}_{(c)}, \mathbf{b}_{(c)}\}$ as the learnable parameters of the critic.

\subsection{Overall Optimization}
Incorporated with the parameterized recommender actor \cref{eq:gm-policy} and the valuation critic \cref{eq:critic}, we can now reformulate the enhanced KL-conservative objective \cref{eq:cdice} as:
\begin{equation}
    \label{eq:dice-kl}
    \begin{aligned}
        & \max_{\pi_\theta} \min_{v_\phi} \log \underset{(\mathbf{s}, \mathbf{a}, \mathbf{s}') \sim d^{\mathcal{D}}}{\mathbb{E}}\left[e^{v_\phi(\mathbf{s}, \mathbf{a}) - \gamma v_\phi(\mathbf{s}', \mathbf{a}')}\right] \\
        & - (1-\gamma)\underset{(\mathbf{s}_0, \mathbf{a}_0) \sim d^{\mathcal{D}}}{\mathbb{E}}\left[v_\phi(\mathbf{s}_0, \mathbf{a}_0)\right] - \mathbb{KL}\left[\pi_{\theta}(\cdot | \mathbf{s}) || \pi_{\theta'}(\cdot | \mathbf{s})\right].
    \end{aligned}
\end{equation}
where $\phi$ denotes the parameters to be optimized in critic network, and $\theta=\{\theta_e, \theta_a\}$ is the parameters of the recommender agent. Since we have no prior knowledge about demonstration set $\mathcal{D}$ for optimization, thus sharing bottom encoding knowledge about user state will not only help reducing parameters otherwise an additional encoder for the critic is needed, but also forming an adversarial competition that emphasis different aspects of state encoding: from one aspect, we would like the recommender agent to imitate what is evaluated high by the critic (the minimization subroutine in \cref{eq:dice-kl}); from another aspect, we would like the recommender agent to reinforce high-evaluated sub-transitions (the maximization subroutine in \cref{eq:dice-kl}) in existing interactive trajectories, but with conservation concerned. Such adversarial knowledge is demonstrated to be useful in previous works \cite{rashidinejad2021bridging,kumar2022should}. \Cref{alg:abcirl} shows training details, where we use second-order Taylor expansion to approximate KL conservation $\mathcal{R}_{\text{kl}}$.
\begin{algorithm}[t]
    \caption{\label{alg:abcirl} Adversarial Batch Conservative iRL}
	\raggedright
	{\bf Input}: compositional demonstration set $\mathcal{D}$. \\
	{\bf Output}: agent $\pi_\theta(\mathbf{a}\mid\mathbf{s})$ and critic $v_\phi(\mathbf{s}, \mathbf{a})$.\\
	\begin{algorithmic} [1]
        \State Initialize parameters $\theta, \phi$.
        \For{$i = 1, \dots, I$}
            \State Sample $\left\{\left(\mathbf{s}^{(b)}_0, \mathbf{s}^{(b)}, \mathbf{a}^{(b)}, \mathbf{s}'^{(b)}\right)\right\}_{b=1}^B \sim \mathcal{D}$
            \State Compute Fisher information matrix $G(\mathbf{s}, \mathbf{a};\theta)$ on $\mathcal{D}$
            \For{iteration $j=1, \dots, B$}
                \State $\mathbf{a}_0^{(j)} \sim \pi_\theta\left(\cdot \mid \mathbf{s}_0^{(j)}\right)$ \Comment{\cref{eq:gm-policy}}
                \State $\mathbf{a}'^{(j)} \sim \pi_\theta\left(\cdot \mid \mathbf{s}'^{(j)}\right)$ \Comment{\cref{eq:gm-policy}}
            \EndFor
            \State $\hat{J}_{log} = \log \left(\frac{1}{B}\sum_{j=1}^{B}\left(e^{v_\phi(\mathbf{s}^{(j)}, \mathbf{a}^{(j)}) - \gamma v_\phi(\mathbf{s}'^{(j)}, \mathbf{a}'^{(j)})}\right)\right)$
            \State $\hat{J}_{linear} = \frac{1}{B}\sum_{j=1}^{B}\left((1-\gamma)v_\phi(\mathbf{s}_0^{(j)}, \mathbf{a}_0^{(j)})\right)$
            \State $\mathcal{R}_{\text{kl}} \approx \frac{1}{B}\sum_{j=1}^{B}\left(\delta \theta^T G(\mathbf{s}^{(j)}, \mathbf{a}^{(j)};\theta) \delta \theta\right)$
            \State Update $\phi \leftarrow \phi - \eta_v \nabla_\phi (\hat{J}_{log} - \hat{J}_{linear})$
            \State Update $\theta \leftarrow \theta + \eta_a \nabla_{\theta}\left(\hat{J}_{log} - \hat{J}_{linear}-\mathcal{R}_{\text{kl}}(\theta)\right)$
        \EndFor 
	\end{algorithmic}
\end{algorithm}
\section{experiments}
In this section, we empirically examine and compare our proposed learning algorithm. We perform experiments on two publicly available real-world datasets, aiming to address the following research questions: (i) \textbf{Effectiveness.} Does adversarial discounted distribution correction \cref{eq:dice-kl} offer more effectiveness compared with other existing methods for interactive recommendation? (ii) \textbf{Efficiency.} Does off-policy evaluation induced by Bellman transformation \cref{eq:bellman} reach the same performance with less demonstration consumption? (iii) \textbf{Adaptivity}. Do other architecture implementations share the same benefaction from incorporating learning objective and conservation designs?

\subsection{Experimental Setup}
\noindent \textbf{Datasets.} We conduct experiments on two real-world interactive recommendation datasets, \ie \textit{Kaggle}\footnote{\url{https://www.kaggle.com/retailrocket/ecommerce-dataset}} and \textit{RecSys15}\footnote{\url{https://recsys.acm.org/recsys15/challenge}}. 
\begin{itemize}
    \item \textbf{Kaggle} This dataset is released by a real-world e-commerce platform and provides a more uniform coverage over interactions, thus is suitable for comparison with reinforcement learning baselines designed for this setting. To align with the \textit{RecSys15}, we consider views as clicks and adding items to the cart as purchases. We remove items interacted fewer than 3 times, as well as interactions smaller than 3, details in \Cref{table:data}.
    \item \textbf{RecSys15} This dataset is released by RecSys Challenge 2015 and provides a more compound coverage over interactions, which offers a setting to compare with imitation learning baselines developed for expert demonstrations. We eliminate interactions smaller than 3 and subsequently sample 200,000 interactions, details in \Cref{table:data}.
\end{itemize}

\begin{table}[t]
  \centering
  \caption{Data Statistics.}
  \setlength{\tabcolsep}{25pt}
  \label{table:data}
  \scalebox{1.0}{
    \begin{tabular}{@{}lll@{}}
      \toprule[1pt]
      & \textit{Kaggle} & \textit{RecSys15} \\
      \midrule
      \#interactions & 195,523 & 200,000 \\
      \#items & 70,852 & 26,702 \\
      \#clicks & 1,176,680 & 1,110,965 \\
      \#purchases & 57,269 & 43,946 \\
      \bottomrule[1pt]
    \end{tabular}
  }
\end{table}

\noindent \textbf{Metrics.} For offline evaluation, we measure top-k $(k=\{5, 10\})$ Hit Ratio (H@k) \cite{xin2020self} and Normalized Discounted Cumulative Gain (N@k) \cite{jarvelin2002cumulated}, which are widely adopted as a measurement for recalling and ranking performance in recent works\cite{xiao2021general,xin2020self}. To ensure that the dataset is divided into non-overlapping subsets for different purposes, we randomly select 80\%  as the training set, 10\% as the validation set, and the remaining interactions as the test set. 

\noindent \textbf{Baselines.} We consider following learning algorithms for comparison: Behavior Cloning (BC) \cite{ross2010efficient} and Supervised Learning (SL) \cite{hidasi2018recurrent}, policy gradient for actor with supervised learning to reward (SL+PG)\cite{gong2019exact}, off-policy Actor-Critic (SL+AC) \cite{xiao2021general}, adversarial policy learning (AL+PG)\cite{bai2019model} and adversarial Deep Q-Learning (AL+DQN)\cite{chen2019generative}.  Specifically, we adopt original settings \cite{xin2020self} for reward-set baselines: 0.2 for click, 1.0 for purchase, and 0.0 for passing as reward-set baselines. A 2-layer GRU with 64 hidden units, is used as backbone for all baselines. We use 10 recent interactions as input length ($w=10$), with mini-batch $B=256$ and state dimensions $d_s = 64$. Item embeddings are initialized from Gaussian distribution ($d_e = 50$). For recommender agent \cref{eq:gm-policy}, we adopt a 2-layer MLP with ReLU as nonlinear activation, the scalar temperature $\gamma_g$ for Gumbel-Softmax is 0.2 and conservative scalar $\delta$ is $1e-2$ as SL+AC \cite{xiao2021general}. We utilize the same MLP for the critic network (\cref{eq:critic}), both with 512 hidden units, and the regression parameter is 3-dim ($l=3$) as a representation of pass, click and purchase feedbacks. Learning rates $\eta_v, \eta_a$ are $5e-3$ for 50 epochs.

\subsection{Experimental Results}
\noindent \textbf{Overall Performance (i).} \Cref{table:click/gru} shows click performance among comparing baselines, and \Cref{table:purchase/gru} gives the results on purchase feedback. Both experiments are conducted on GRU backbone. A similar tendency can be observed in both tables. First, we observe that BC works worst among these baselines on both datasets, which demonstrates that vanilla imitation learning does not suit compositional demonstrations in interactive recommendation, and RL-based IRS can reveal new valuable patterns even in offline environments, same as previous work reports \cite{rashidinejad2021bridging}. Second, we observe that off-policy methods (SL+AC and AL+DQN) work better than on-policy methods (SL+PG and AL+PG) in either model-based or model-free groups, because on-policy methods generally acquire online interactions to evaluate current agent while this is not available in offline environments. Third, we observe that model-based methods (AL+PG) works better than model-free approach (SL+PG) on more compositional demonstrations, \ie \textit{RecSys}, and vice versa for more uniform demonstrations, \ie \textit{Kaggle}, this is consistent with existing works \cite{rashidinejad2021bridging}. Finally, our proposed method outperforms  all compared learning algorithms, which results from the combination between learning to reward (the minimization in \cref{eq:dice-kl}) and policy reinforcement (the maximization in \cref{eq:dice-kl}).
\begin{table}[t]
  \centering
  \caption{Effectiveness. Best is bold, and the next best is underlined. ``{\bf $\ast$}'' indicates the statistically significant improvements (two-sided t-test with $p<0.05$) over the best baseline.}
  \label{table:click/gru}
  \scalebox{.7}{
    \begin{tabular}{@{}lllllllll@{}}
      \toprule[1pt]
      \multirow{ 2}{*}{click} & \multicolumn{4}{c}{RecSys} & \multicolumn{4}{c}{Kaggle} \\ \cmidrule(lr){2-5} \cmidrule(lr){6-9}
      & H@5 & N@5 & H@10 & N@10 & H@5 & N@5 & H@10 & N@10 \\
      \midrule
      BC & .2107 & .1264 & .3179 & .1628 & .1288 & .1134 & .1784 & .1332 \\
      SL & .2876& .1982& .3793& .2279& .2233& .1735& .2673 & .1878 \\
      \midrule
      SL+PG & .3012 & .2106 & .4013 & .2382 & .2504 & .1972 & .3036 & .2118 \\
      SL+AC & \underline{.3276} & \underline{.2306} & \underline{.4217} & \underline{.2593} & \underline{.2659} & .2181 & .3204 & .2351 \\
      \midrule
      AL+PG & .3034 & .2084 & .4022 & .2351 & .2589 & .2053 & .3142 & .2189 \\
      AL+DQN & .3249 & .2271 & .4208 & .2583 & .2658 & \underline{.2289} & \underline{.3263} & \underline{.2478} \\
      \midrule
      Our & \best{.3314} & \best{.2572} & \best{.4459} & \best{.2712} & \best{.2750} & \best{.2431} & \best{.3363} & \best{.2647} \\
      \bottomrule[1pt]
    \end{tabular}
  }
\end{table}

\begin{table}[t]
  \centering
  \caption{Effectiveness. Best is bold, and the next best is underlined. ``{\bf $\ast$}'' indicates the statistically significant improvements (two-sided t-test with $p<0.05$) over the best baseline.}
  \label{table:purchase/gru}
  \scalebox{.7}{
    \begin{tabular}{@{}lllllllll@{}}
      \toprule[1pt]
      \multirow{ 2}{*}{purchase} & \multicolumn{4}{c}{RecSys} & \multicolumn{4}{c}{Kaggle} \\ \cmidrule(lr){2-5} \cmidrule(lr){6-9}
      & H@5 & N@5 & H@10 & N@10 & H@5 & N@5 & H@10 & N@10 \\
      \midrule
      BC & .2772 & .1758 & .4142 & .2338 & .1939 & .1618 & .3379 & .1821 \\
      SL & .3994 & .2824 & .5183 & .3204 & .4608 & .3834 & .5107 & .3995 \\
      \midrule
      SL+PG & .4325 & .3071 & .5412 & .3414 & .5087 & .4172 & .5602 & .4340 \\
      SL+AC & \underline{.4427} & \underline{.3219} & \underline{.5571} & \underline{.3587} & .5341 & .4339 & .5868 & .4687 \\
      \midrule
      AL+PG & .4204 & .3041 & .5394 & .3360 & .5158 & .4328 & .5724 & .4577 \\
      AL+DQN & .4353 & .3183 & .5415 & .3545 & \underline{.5374} & \underline{.4383} & \underline{.5894} & \underline{.4719} \\
      \midrule
      Our & \best{.4452} & \best{.3259} & \best{.5637} & \best{.3686} & \best{.5459} & \best{.4460} & \best{.5961} & \best{.4739} \\
      \bottomrule[1pt]
    \end{tabular}
  }
\end{table}
\noindent \textbf{Efficiency Study (ii).} \Cref{table:efficiency} shows the efficiency comparison among RL algorithms in \textit{RecSys} on GRU backbone. We use SL performance in \cref{table:click/gru} as the threshold, and count iterations needed for the agent to continuously exceed SL in 5 times as a measurement for exploration-efficiency. Epochs averaged over 10 experiments are reported as results. First, we observe that both off-policy approaches exceed on-policy methods, either model-based or model-free, this verifies the motivation to develop an off-policy version of learning objective \cref{eq:dice}. Next, we observe that adversarial learning (AL+PG and AL+DQN) requires more epochs than supervised learning (SL+AC and SL+PG), since the dynamic equilibrium of the former generally requires more time to fit. Our approach requires the least epochs (relative 11.53\% reduction), because Bellman transformation \cref{eq:bellman} results in off-policy evaluation, and the objective \cref{eq:dice-kl} unifies reinforcement learning and auxiliary learning (AL or SL) to reduce complexity.

\begin{table}[t]
  \centering
  \caption{Efficiency}
  \setlength{\tabcolsep}{18pt}
  \label{table:efficiency}
  \scalebox{1.0}{
    \begin{tabular}{@{}llll@{}}
      \toprule[1pt]
      & model & policy & efficiency \\
      \midrule
      SL+AC\cite{xiao2021general} & free & off & 10.4 ($\pm$ 0.49) \\
      SL+PG\cite{gong2019exact} & free & on & 12.3 ($\pm$ 0.48) \\
      \midrule
      AL+PG\cite{bai2019model} & based & on & 13.5 ($\pm$ 0.51) \\
      AL+DQN\cite{chen2019generative} & based & off &  12.4 ($\pm$ 0.66) \\
      \midrule
      \textbf{Ours} & free & off & 9.2 ($\pm$ 0.87) \\
      \bottomrule[1pt]
    \end{tabular}
  }
\end{table}

\begin{figure}[htbp]
    \begin{subfigure}{.49\linewidth}
        \centering
        \includegraphics[height=0.15\textheight]{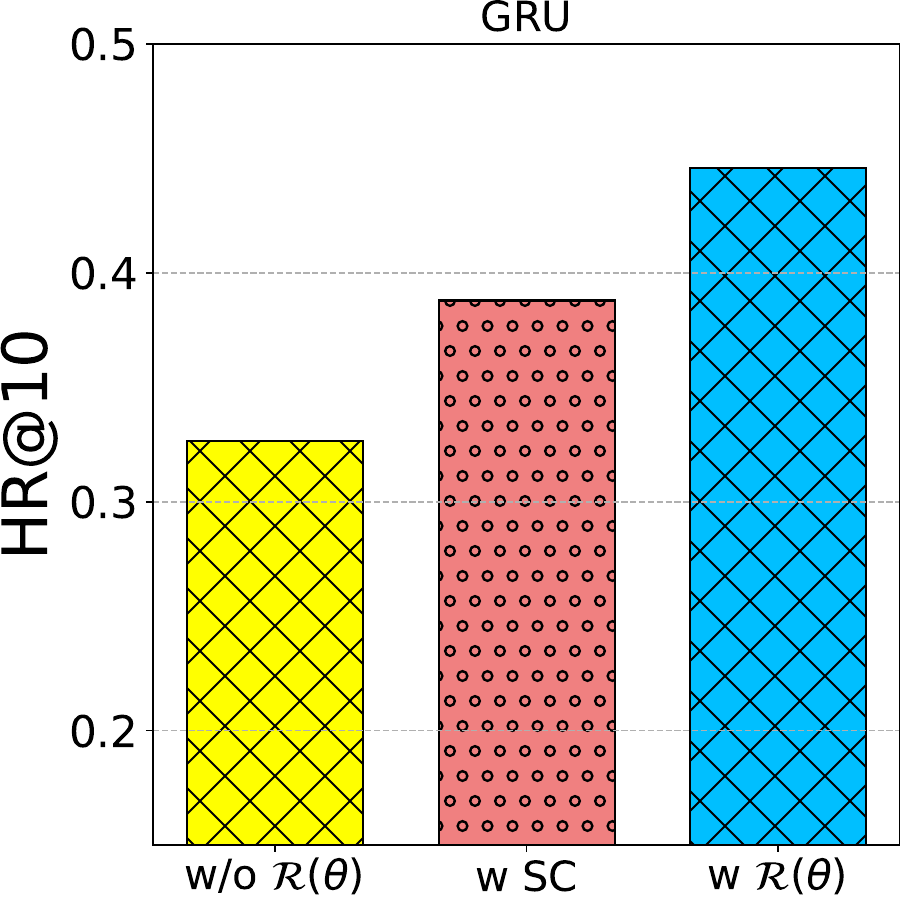}
        \caption{}
        \label{fig:gru-h@10}
    \end{subfigure}
    \hfill
    \begin{subfigure}{.49\linewidth}
        \centering
        \includegraphics[height=0.15\textheight]{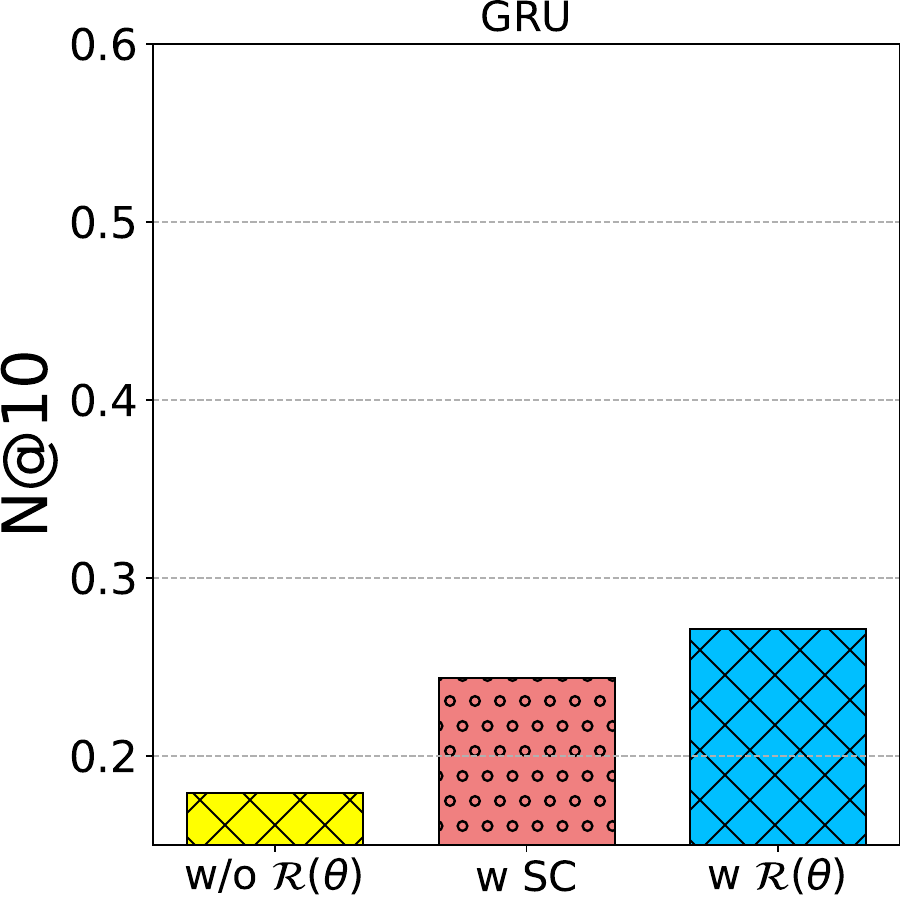}
        \caption{}
        \label{fig:gru-n@10}
    \end{subfigure}
    \\
    \begin{subfigure}{.49\linewidth}
        \centering
        \includegraphics[height=0.15\textheight]{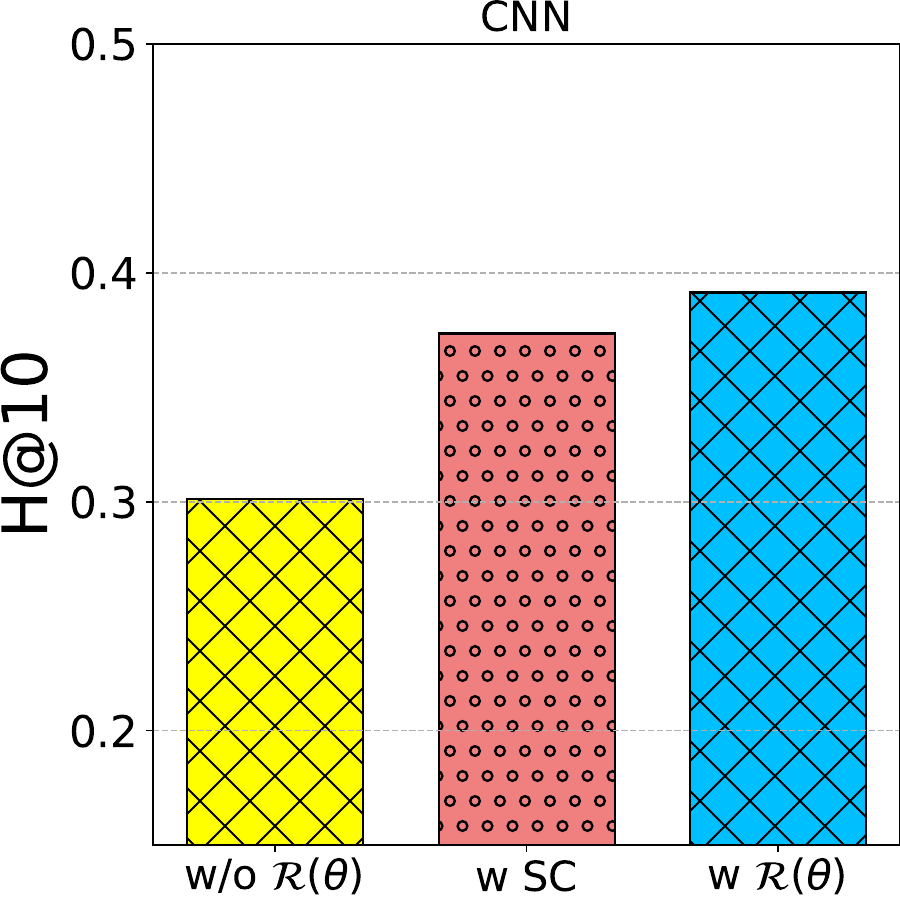}
        \caption{}
        \label{fig:cnn-h@10}
    \end{subfigure}
    \hfill
    \begin{subfigure}{.49\linewidth}
        \centering
        \includegraphics[height=0.15\textheight]{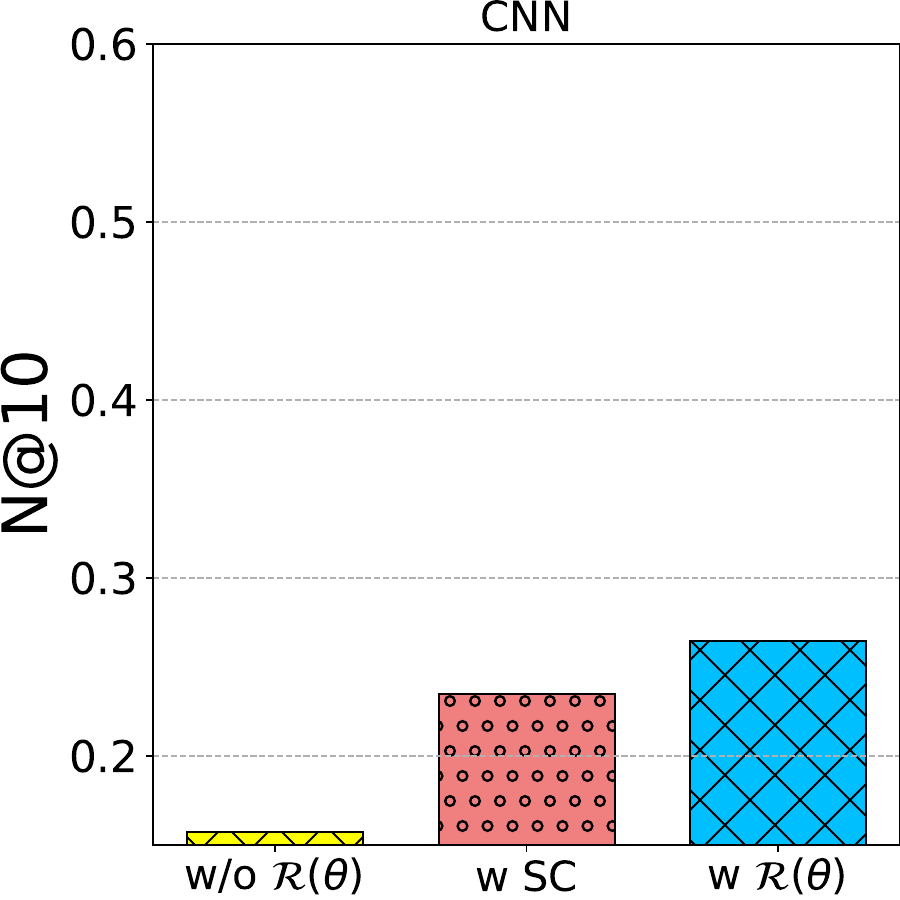}
        \caption{}
        \label{fig:cnn-n@10}
    \end{subfigure}
    \caption{Adaptivity study on click among two backbones.}
    \label{fig:adaptivity}
    \vspace{-4mm}
\end{figure}

\noindent \textbf{Adaptivity analysis (iii).} \Cref{fig:adaptivity} shows the ablation study on \textit{RecSys} with two kind of backbones: GRU and CNN \cite{xiao2021general}\cite{xin2020self}. For the latter, we concatenate input interactions to form a 2D feature map and then conduct convolution upon it. We also implement support constraints \cite{xiao2021general} as a simplified version of conservation. \Cref{fig:gru-h@10} and \Cref{fig:cnn-h@10} show results on H@10, \Cref{fig:gru-n@10} and \Cref{fig:cnn-n@10} show results on N@10. Vanilla objective ($w/o\ \mathcal{R}(\theta)$) performs close to SL, since offline demonstrations do not cover all items for the agent to explore. Uncertainty necessitates regularization. SC performs a simplified conservation from a supervised-learnt behavior agent. Since behavior agent estimation has inaccuracy, direct conservation ($w\ \mathcal{R}(\theta)$) achieves best improvement.

\section{Related Works}

Classic recommendation algorithms \ie \cite{rendle2010factorizing} assume that similar users have similar preferences and propose collaborative filtering algorithms based on matrix factorization. However, classic methods cannot effectively model high-order user-agent interaction dynamics. To address this issue, deep sequential recommendation approaches, \ie \cite{hidasi2018recurrent} treat interaction procedures as temporal sequences, and use latent state vectors to capture the high-order temporal dynamics of user preferences. But in interactive recommendation tasks, there are multi-types feedback signals with different valuations for the RA, \eg user click behavior may better reflect their true interests than purchase behavior. Deep sequential models do not contain this difference when modeling user behavior.

To further address this, RL-based recommender agent aims to optimize the cumulative reward function from various feedback signals, existing works follow as: (i) \textbf{policy-based methods}, considering the constraints of real-time user interactions in recommendation problems, off-policy REINFOCE\cite{chen2019top,gong2019exact} uses a reweighting method based on propensity scores for video recommendations on the YouTube platform. (ii) \textbf{value-based methods}, SQN \cite{xin2020self,chen2019generative} utilizes temporal difference learning to learn the maximization of cumulative value rewards, and jointly minimizes cross-entropy temporal predictions to capture preference tendency. (iii) \textbf{actor-critic methods}. SDAC\cite{xiao2021general} proposes a policy estimation model based on the Gumbel distribution to address the discretization of the action space. However, heuristic-designed reward functions require burdensome fine-tuning to ensure the stability of the reinforcement learning training process. In this work, we avoid reward fine-tuning by learning to reward from offline demonstrations, which is more effective and adaptive.
\section{conclusion}
In this work, we propose a novel batch inverse reinforcement learning algorithm for interactive recommendation. We combine learning-to-reward procedure and off-policy evaluation with a unified discounted distribution correction objective, and impose conservative KL penalty upon a vanilla objective since offline interactive demonstrations can be compositional without prior knowledge. Empirical studies on two real-world datasets justify the effectiveness and efficiency of our proposed methods, and further adaptivity analysis confirms that our solution is applicable to other neural architectures. While current algorithm relies purely on offline demonstrations, a mixture of offline demonstrations and online interaction will be explored in the future.

\section*{acknowledgement}
This research is supported by APRC - CityU New Research Initiatives (No.9610565, No.9360163), Hong Kong ITC Fund Project (No.ITS/034/22MS), and SIRG - CityU Strategic Research Grant (No.7020046, No.7020074, No.7005894).

\bibliographystyle{IEEEtran}
\bibliography{IEEEabrv, 7_simplified_references}

\end{document}